\documentclass[conference]{IEEEtran}
\usepackage{cite}
\usepackage{amsmath,amssymb,amsfonts}
\usepackage{graphicx}
\usepackage{textcomp}
\usepackage{xcolor}
\usepackage{tabularx}
\usepackage[hidelinks]{hyperref}
\usepackage{url} 
\begin{document}

\title{15,500 Seconds: Lean UAV Classification Using EfficientNet and Lightweight Fine-Tuning}

\author{\IEEEauthorblockN{Andrew P. Berg}
\IEEEauthorblockA{\textit{Department of Computer Science} \\
\textit{College of Charleston}\\
Charleston, SC, USA\\
berga2@g.cofc.edu}
\and
\IEEEauthorblockN{Qian Zhang}
\IEEEauthorblockA{\textit{Department of Engineering} \\
\textit{College of Charleston}\\
Charleston, SC, USA \\
zhangq@cofc.edu}
\and
\IEEEauthorblockN{Mia Y. Wang}
\IEEEauthorblockA{\textit{Department of Computer Science} \\
\textit{College of Charleston}\\
Charleston, SC, USA \\
wangy5@cofc.edu}

}

\maketitle

\begin{abstract}
As unmanned aerial vehicles (UAVs) become increasingly prevalent in both consumer and defense applications, the need for reliable, modality-specific classification systems grows in urgency. This paper addresses the challenge of data scarcity in UAV audio classification by expanding on prior work through the integration of pre-trained deep learning models, parameter-efficient fine-tuning (PEFT) strategies, and targeted data augmentation techniques. Using a custom dataset of 3,100 UAV audio clips (15,500 seconds) spanning 31 distinct drone types, we evaluate the performance of transformer-based and convolutional neural network (CNN) architectures under various fine-tuning configurations. Experiments were conducted with five-fold cross-validation, assessing accuracy, training efficiency, and robustness. Results show that full fine-tuning of the EfficientNet-B0 model with three augmentations achieved the highest validation accuracy (95.95\%), outperforming both the custom CNN and transformer-based models like AST. These findings suggest that combining lightweight architectures with PEFT and well-chosen augmentations provides an effective strategy for UAV audio classification on limited datasets. Future work will extend this framework to multimodal UAV classification using visual and radar telemetry.
\end{abstract}


\section{Introduction}
Unmanned Aerial Vehicles (UAVs), or commonly known as drones, pose security and intelligence threats to private and public interests. There is a growing need in defense and intelligence sectors for UAV Classification as modern methods and systems rely on less-robust UAV detection. UAV detection is not enough, as more drones flood the airs.

There are a variety of ways to classify UAVs using various modalities: Radio frequency (RF), Radar, visual, and audio. In this work we focus on UAV audio classification as a starting point in future works we aim to improve the scope incorporating multiple modalities. Further, because of the scale and complexity of our data we have opted to use deep learning (DL) neural networks.

A key issue in all of the listed modalities is data scarcity. It is relatively expensive to capture the necessary telemetry for all available UAVs to classify. In this paper we build on our lab's previous work to solve this problem using data, and architecture training regimes such as: parameter efficient fine-tuning, data augmentation, and pre-trained networks. 

In the following sections we will discuss previous work and literature that encouraged our findings. A deep dive into our methodology, including: how our dataset was collected, training environment, and a breakdown of the solutions used to train for classification networks. Lastly, we will provide an empirical analysis with 5-fold cross validated results, where we achieve upwards of 95\% validation accuracy using EfficientNet.

\textbf{The projects research questions are as follows:}
\begin{enumerate}
    \item On our custom UAV dataset can fine-tuned transformers outperform fine-tuned CNNs? 
    \item What is the optimal combination of PEFT, data augmentations, and model selection to achieve the highest result?
\end{enumerate}

\section{Literature Review}

\textbf{Previous Work:} This paper is primarily based on the previous work done by our lab in \cite{b1}. The prior paper explored an extremely small data approach of a 9-class dataset (4,500 seconds) and UAV audio classification; comparing transformers and CNNs. The former paper concluded that our custom CNN approach had better performance. Importantly, when working with deep learning models such as CNNs and transformers what we deem small data is entirely relative to the model architecture. Our custom dataset is large and complex enough for a less complex machine learning (ML) system, whereas the neural networks are more data hungry.

\textbf{Audio Classification:} We use deep audio classification to refer to audio classification performed using neural networks. The study \cite{b2}. Found the most common approach to involve CNNs, using feature extraction techniques to transform the raw audio content using various methods like relative waves and mel-scale transformations. Another paper  \cite{b3} compares pre-trained popular transformers and CNNs against one another further motivating our work, and guiding our model selections.

\textbf{UAV Classification:} Reference \cite{b48} breaks down the various modalities and their drawbacks. Further it explores the different machine and deep learning approaches to the problem. Other studies look into the various open UAV datasets \cite{b48}and approaches and the different approaches taken to them.

\section{Methodology}
\subsection{Data Collection}
The drone data collection is still an on-going project \cite{b50} \cite{b51}. So far, for the past 5 years, drone audio data was collected from 31 distinct unmanned aerial vehicles (UAVs), as detailed in Table~\ref{tab:uav_data_31}. Each UAV contributed 100 five-second audio recordings, resulting in a total of 3,300 audio files and 15,500 seconds of raw flight audio. The dataset is perfectly balanced. These recordings encompass a broad spectrum of consumer, commercial, and custom-built UAV platforms and were conducted across both indoor and outdoor environments in three U.S. locations.

\textbf{Drone Overview:} The dataset includes 28 quadcopters, one tricopter, and one hexacopter. Most of the quadcopters use a conventional X-frame design typical of consumer drone platforms. UAVs were sourced from manufacturers including DJI, Autel, Syma, Yuneec, UDI, Hasakee, Holystone, Hover, and two self-built drones. Notably, \textit{David Tricopter}, designed by David Windestal, is a custom-built tricopter with a 34-inch diameter, equipped with an AfroFlight Naze32 flight controller and weighing approximately 2.6 lbs. \textit{PhenoBee}, designed by Ziling Chen, is the largest UAV in the dataset, weighing approximately 23 kg with a 1.35-meter diameter, and operates using the Ardupilot framework with Cube Orange hardware.

\textbf{Recording Sites:} Data were collected in West Lafayette, Indiana; New Richmond, Indiana; and Charleston, South Carolina. Indoor recordings in Indiana were conducted in a university lab setting, while outdoor recordings were collected on a private farm in New Richmond. Charleston indoor data were collected in the Drone Lab at the Harbor Walk Campus of the College of Charleston, and outdoor recordings were captured from the rooftop of the aquarium parking garage. Environmental factors varied naturally during recordings and included wind, birdsong, traffic noise, and changing weather conditions.

\textbf{Recording Equipment:} From 2021 through 2023, audio data were captured using a MacBook Air with a 1.1GHz quad-core Intel Core i5 processor and 8GB of memory. Beginning in 2024, recordings used a MacBook Air with an Apple M3 chip and 16GB of memory. All data were recorded using the system's built-in microphones without external post-processing.

\begin{table*}[ht]
\centering
\caption{UAV Audio Dataset: 31 Classes with Collection Sites}
\label{tab:uav_data_31}
\begin{tabular}{lllrrl}
\hline
\textbf{Manufacture} & \textbf{Model} & \textbf{Drone Type} & \textbf{Number of Files} & \textbf{Duration (sec)} & \textbf{Collection Site} \\
\hline
Self-build & David Tricopter & Outdoor & 100 & 500 & Columbus, IN \\
Self-build & PhenoBee & Outdoor & 100 & 500 & West Lafayette, IN \\
Autel & Evo 2 Pro & Outdoor & 100 & 500 & New Richmond, IN \\
DJI & Avata & Outdoor & 100 & 500 & Charleston, SC \\
DJI & FPV & Outdoor & 100 & 500 & Charleston, SC \\
DJI & Matrice 200 & Outdoor & 100 & 500 & West Lafayette, IN \\
DJI & Matrice 200 V2 & Outdoor & 100 & 500 & New Richmond, IN \\
DJI & Matrice 600p & Outdoor & 100 & 500 & New Richmond, IN \\
DJI & Mavic Air 2 & Outdoor & 100 & 500 & New Richmond, IN \\
DJI & Mavic Mini 1 & Outdoor & 100 & 500 & New Richmond, IN \\
DJI & Mini 2 & Outdoor & 100 & 500 & New Richmond, IN \\
DJI & Mini 3 & Outdoor & 100 & 500 & Charleston, SC \\
DJI & Mini 3 Pro & Outdoor & 100 & 500 & Charleston, SC \\
DJI & Mavic 2 Pro & Outdoor & 100 & 500 & New Richmond, IN \\
DJI & Mavic 2s & Outdoor & 100 & 500 & New Richmond, IN \\
DJI & Phantom 2 & Outdoor & 100 & 500 & New Richmond, IN \\
DJI & Phantom 4 & Outdoor & 100 & 500 & New Richmond, IN \\
DJI & Tello & Indoor & 100 & 500 & Charleston, SC \\
DJI & RoboMaster TT Tello & Indoor & 100 & 500 & New Richmond, IN \\
Hasakee & Q11 & Indoor & 100 & 500 & West Lafayette, IN \\
Holystone & HS210 & Indoor & 100 & 500 & Charleston, SC \\
Hover & X1 & Outdoor & 100 & 500 & Charleston, SC \\
Syma & X5SW & Indoor & 100 & 500 & West Lafayette, IN \\
Syma & X5UW & Indoor & 100 & 500 & West Lafayette, IN \\
Syma & X8SW & Indoor & 100 & 500 & West Lafayette, IN \\
Syma & X20 & Indoor & 100 & 500 & West Lafayette, IN \\
Syma & X20P & Indoor & 100 & 500 & West Lafayette, IN \\
Syma & X26 & Indoor & 100 & 500 & West Lafayette, IN \\
Swellpro & Splash 3 plus & Outdoor & 100 & 500 & New Richmond, IN \\
Yuneec & Typhoon H Plus & Outdoor & 100 & 500 & New Richmond, IN \\
UDI RC & U46 & Outdoor & 100 & 500 & West Lafayette, IN \\
\hline
 & & Total & 3,100 & 15,500\\
\hline
\end{tabular}
\end{table*}

\subsection{Technical Environment}
\textbf{Software:}
The code is available at our public Github repository \cite{b4}
The model training runs are tracked using Weights \& Biases and are publicly available at \cite{b5}

This project utilized various machine learning frameworks. We list the following with brief descriptions and references: Pytorch as the essential deep learning framework of chioce\cite{b6}\cite{b7}; Transformers and PEFT packages\cite{b8}\cite{b9}; Pytorch lighting to simplify training loops \cite{b10}; Weights and Biases for experiment logging and hyperparameter searching \cite{b11}; Docker for server and version management \cite{b12}. Other python packages such as: Torchaudio \cite{b13}, Audiomentations\cite{b14}, TorchMetrics\cite{b15}, MatPlotLib \cite{b16}, SciKitlearn\cite{b17}, NumPy\cite{b18}, and Librosa\cite{b19} were used.

\textbf{Hardware:}
We ran all of our model experiments on a local server built with an AMD Threadripper 4950x, 128 Gbs of RAM, and two NVIDIA GeForce RTX 4090s; both equipped with 24 Gbs of VRAM. Notably, we chose to only train with one of the two GPUs. This is because the distributed runs took longer. In future work this may be a bottleneck; however, for our current data scale it is not prohibitive.

\subsection{Feature Extraction}
\begin{figure}[ht]
    \centering
   
     \caption{Audio Analysis of DJI Tello Drone}
    \label{fig:audio-analysis}
    \includegraphics[width=0.4\textwidth]{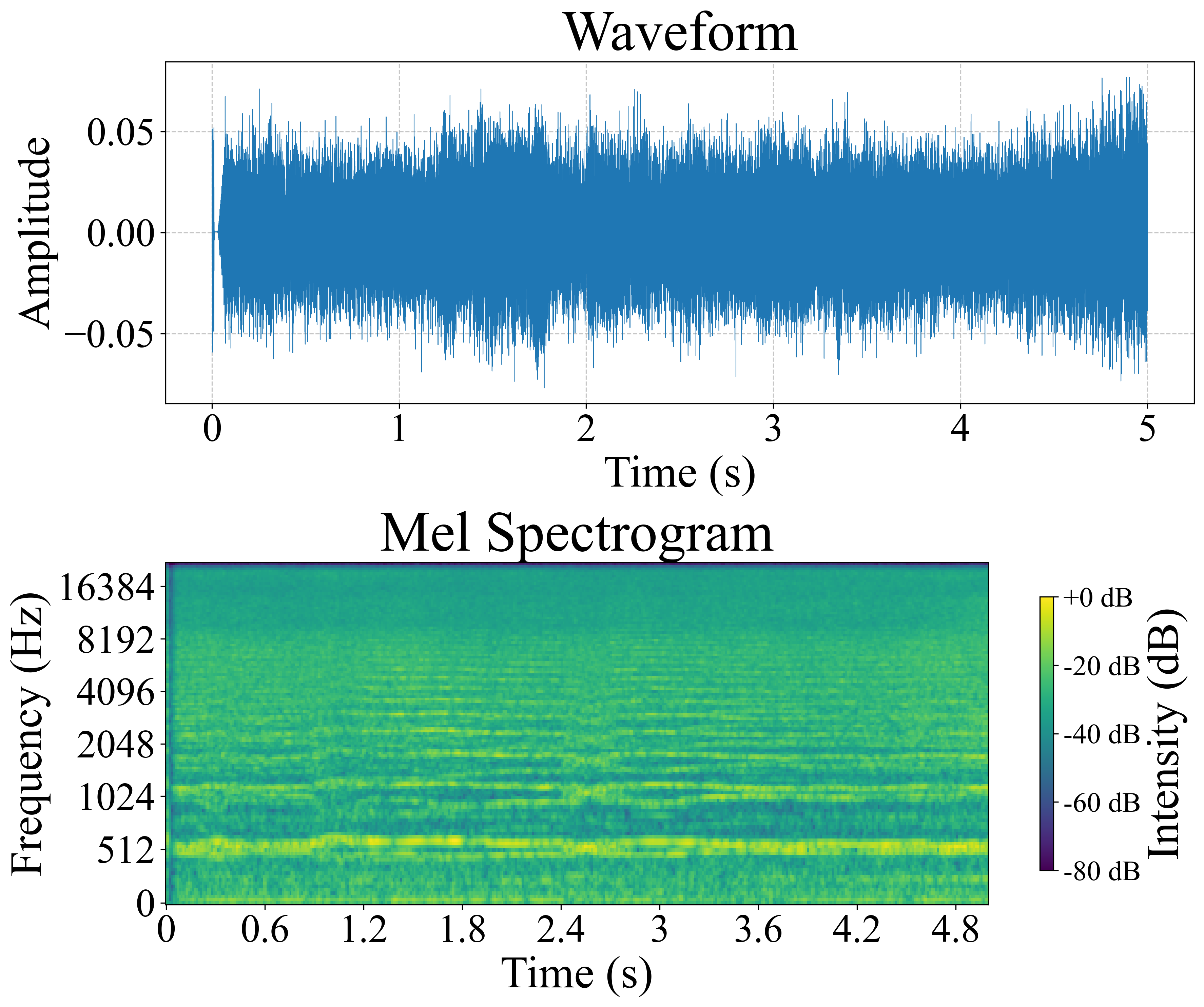}
\end{figure}

The raw recorded UAV audio is created in waveform format. This representation of a sound signal is a time-domain signal that represents how loud a sound (amplitude) is over time and is commonly used for playback to human ears. The spectrogram, however, is a visual representation of how the frequency content changes over time. Approximated into small windows. The spectrogram leverages the fourier transform for each window. A mel-spectrogram produces a visual representation of sound adjusted for human perception. Both wave and the mel-spectrogram of a raw drone sample are shown in figure \ref{fig:audio-analysis}.

Feature extraction is a staple in Audio Classification \cite{b21} and greatly improves the performance of models in audio tasks. Further, there is precedent for using mel-spectrograms as used in previous UAV audio classification \cite{b22}.

\subsection{Data Augmentations}
Data augmentations vary in their application and methods. We leverage empirical literature \cite{b23} to narrow our scope of relevant UAV audio augmentations. We chose to perform the data augmentations on the raw audio content, then feature extract. Further, we do not perform any augmentations on the spectrogram data. Please note that the augmented samples inflate the training set; meaning, instead of replacing the original samples.

The usage of data augmentations addresses the inherit data scarcity. In previous work we tested and found the optimal combination and configurations for the data augmentations \cite{b1} improving certain model training performances. We found the most relevant augmentations to be time stretch combined with sin distortion. Optimally we use 3 augmentations per training sample. Note that we don’t inflate the test set, to maintain generalization.

However, since we scaled up from our previous work the problem of data scarcity may no longer be largest limiting factor to model’s performance. For the full breakdown and visualization of the augmentations used please refer to the colab notebook \cite{b24}

\subsection{Training}

In this paper we utilize the pre-training/fine-tuning paradigm of model training. We define a pre-trained model as any model that has been trained on a dataset or task before we train it. We leverage these pre-trained models using various fine-tuning strategies to effectively learn for a new smaller dataset and downstream task\cite{b25}. Essentially we are re-purposing the model’s learned representations and applying it to new data; in this case UAV audio classification.

For our training regime we employ cross entropy loss\cite{b26}, and the Adam optimizer\cite{b27}. Notably, for the transformer model we use AdamW \cite{b28} optimizer with the weight decay set to 0.01. We train at a fixed learning rate of 0.001 and reduce it on plateaus by a scaling factor of 0.1. We use a fixed batch size of 8, and the accumulation steps set to 2. Effectively simulating a batch size of 16. We split the normal training runs into a train, test, validation, inference split set to 60,20,10,10 respectively. Both the training and validation splits have augmented data, whereas the validation and inference splits do not. This split allows us to see our model's generalizability and avoid the risks of overfitting to the training set.

\textbf{Full Fine-tuning:} The most common way to adapt a pre-trained model to a dataset is to train all of the model’s parameters. This is feasible for large enough datasets, but is slow and fails and overwrites and has massive forgetting from the pre-trained weights

\textbf{Linear Probing:} Linear probing, or as we refer to it in our experiments, classifier fine-tuning, leaves only the classifier block trainable \cite{b29}. This is an ablation study tactic used to judge a model's pre-trained generalization to a new task. This fine-tuning technique isn't reliable as it doesn’t have the effective learned representations to generalize effectively and is included primarily as a benchmark against other methods of fine-tuning. 

\subsection{PEFT}
Parameter efficient Fine tuning is an general term that describes any training method that leverages pre-trained neural network that only trains on a small subset of a model's total parameter count. We draw much inspiration and instruction from \cite{b48}. We differentiate between the PEFT methods that can be used with transformers and those that can’t.

\textbf{OFT (Orthogonal Fine Tuning):} Initially proposed for the fine-tuning of text to image generation models, OFT aims to capture what \cite{b30} refers to as “hyperspherical energy” of a model. This boils down to the higher importance of a parameter’s orthogonal (diagonal) values. OFT does this by multiplying a model’s weights by a scaled and rotated orthogonal weight. We implemented this PEFT method from the PEFT package from Hugging Face. In theory this could be implemented for both CNNs and Transformers. We however opted to use it on its intended structure of transformers.

\textbf{Ia3 (Infused Adapter by Inhibiting and Amplifying Inner Activations):} Proposed to rival in-context learning in large language models (LLMs) \cite{b31}, Ia3 introduces new learnable vectors that target the query, key, value attention mechanism as well as after the non-linear activation function after the attention mechanism. We implemented this PEFT method from the PEFT package from Hugging Face. This PEFT method is only usable with transformers as it targets the attention mechanism.

\textbf{SSF (Scale Shift Fine Tuning):} SSF was proposed to rival the more computationally expensive PEFT methods \cite{b32}. This PEFT method is purely additive, meaning that it simply introduces a learnable scaling and additive vector to the targeted weights. We chose to implement this PEFT method from scratch and have included it in our repo, its code can be found directly in our repository at \cite{b33}. This PEFT method is model agnostic and we use it for both CNN and Transformer models.

\textbf{Batchnorm Fine-tuning:} Batchnorm Fine-tuning is a selective PEFT method that was initially proposed in \cite{b34}, which showcases solid performance whilst only training on batchnorm layers in CNNs. We implemented this PEFT method ourselves, simply only training all batchnorm parameters. Of course, we also keep the classifier as trainable. This PEFT method is only viable for our CNN models as transformers do not use batchnorm layers. Instead transformers opt for the more parallelizable and stable LayerNorm layers.

\subsection{Models} A key motivating question is: are Transformers or CNNs better for our use case? The main delimiter, we find, is the scale of the data. Dually, the indicative bias of the convolutional network has to image and dually audio modalities is very effective in small data tasks like our own. However, when trained on a high enough quantity of data the transformer can surpass CNN performance on key benchmarks. This is noted in the vision transformer white paper \cite{b35} which explores a purely attention based image classification network.

We can partially overcome the issue of lack of data by using pre-trained models/networks that already have baked in “learned representations” of data. Essentially reusing already learned data towards a new end task. Despite it being a different modality the pre-training/fine-tuning paradigm is empirically proven to be more effective than random initialization(s) \cite{b36}

\textbf{AST (Audio Spectrogram Transformer)} Audio Spectrogram Transformer (AST) is a pre-trained Transformer, which is based on the ViT model, integrating and merging the pre-trained ViT b-16 weights into its own model. Further it is trained on the ImageNet and Audioset datasets. It is a stark example of both pure attention based audio classification and cross modality transfer learning, using both image and audio data to train on.

The Audio Spectrogram Transformer (AST) \cite{b37} is audio classification transformer pre-trained on the ImageNet \cite{b38} and AudioSet \cite{b39} Dataset, leveraging cross modal transfer learning. AST adapts its model architecture from the popular vision transformer (ViT) \cite{b35}.

\begin{figure}[ht]
    \centering
  
    \label{fig:ast}
    \includegraphics[width=0.48\textwidth]{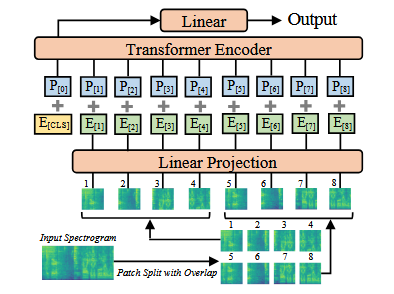}
      \caption{AST visualization. Photo taken from \cite{b37}}
\end{figure}

As shown in figure \ref{fig:ast} the AST uses patchified spectrograms as inputs to the encoder. Notably, there is no decoder structure as seen in natural language processing (NLP) transformers, but instead a fully connected linear layer for classification. We implemented a pre-trained version of AST using the MIT/ast-finetuned-audioset-10-10-0.4593 checkpoint from huggingface. This version was only trained using Audioset.

\begin{figure}[ht]
    \centering
    
    \caption{Custom CNN Architecture}
    \label{fig:custom-cnn}
    \includegraphics[width=0.4\textwidth]{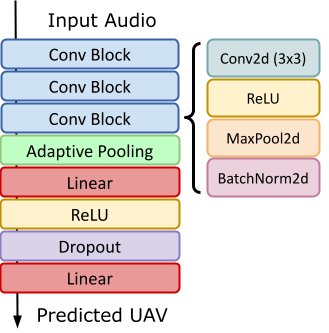}
\end{figure}

\textbf{Custom CNN:} Previously used and tested in \cite{b1}, our standard CNN, pictured in figure \ref{fig:custom-cnn}. Uses 3 stacked convolutional layers with a final fully connected sequence to act as the classifier. This model is not pretrained on any data. Our custom CNN acts as a point of comparison against the other pre-trained models in this paper. 

The model's architecture is as follows: mel-spectrogram input shape into 3 convolutional blocks consisting of conv2d operation with a kernel size of 3 and padding of 1, a ReLU activation function, max average pooling(2D) and a batch normalization(2D) layer. After that a simple dense classification layer that consists of adaptive pooling, a fully connected layer dynamically calculated based on the feature size, a dropout layer, and finally a classifying linear layer set to an output shape of 31.

\textbf{ResNet:} Introduced in \cite{b40} ResNet introduced the residual block into the deep learning field, completely revolutionizing how deep models can be stacked. Simply, the residual block, or skip connection, re-adds a model’s input to a learned weight, this innovation avoids massive forgetting on large/deep neural networks. When it was published, ResNet achieved State of the art (SOTA) performance on the ImageNet benchmark. We implement ResNet v1.5 available at \cite{b41} experimenting on two model sizes: resnet18 and resnet152. We have slightly tweaked the model's layers to agree with our input and output shapes, replacing the first convolutional layer to accept a grayscale input shape. 

\textbf{MobileNet} Originally introduced in 2017, MobileNet \cite{b42}, didn’t aim for the top performance on ImageNet, but rather for efficiency and for the smallest model possible. This aligns with UAV classification applications, which can be applied on edge/mobile devices. We implement the pre-trained updated MobileNet v3 \cite{b43} using both the large and small model variants, sourced from TorchVision \cite{b44}.

\textbf{EfficientNet:} Introduced in 2019 \cite{b45} EfficientNet aimed to strike a balance between a model's scaled depth, width, and depth. In finding an optimal balance between the scaling of CNNs, EfficientNet sits between the extremes of the efficiency of MobileNet and the larger models. We 
Implemented \cite{b46} using the base 0 and base 7 sizes.

\section{Experiments \& Results}

\subsection{Metrics}
We use the following metrics to measure model performance: loss, accuracy, F1 score, loss, training time, and trainable parameter percentage. These metrics are measured across the different data splits (e.g. train loss, validation accuracy, etc.) and all of the models. Not all metrics are visualized, but are available in our logs \cite{b5}

\subsection{Hyperparameter Exploration}
We first started exploring the model's performance by running simple hyperparameter tuning runs. Exploring different configurations for the PEFT methods and models. We use Weight's \& Biases Sweeps tooling to effectively track and visualize the runs.

We used this hyperparameter sweeps to gain intuition on how these models run, and generally what combination of hyperparameter lead to higher performance for each model.

\begin{figure}[ht]
    \centering
   
     \caption{SSF Sweeps of Resnet152}
    \label{fig:sweep}
    \includegraphics[width=0.5\textwidth]{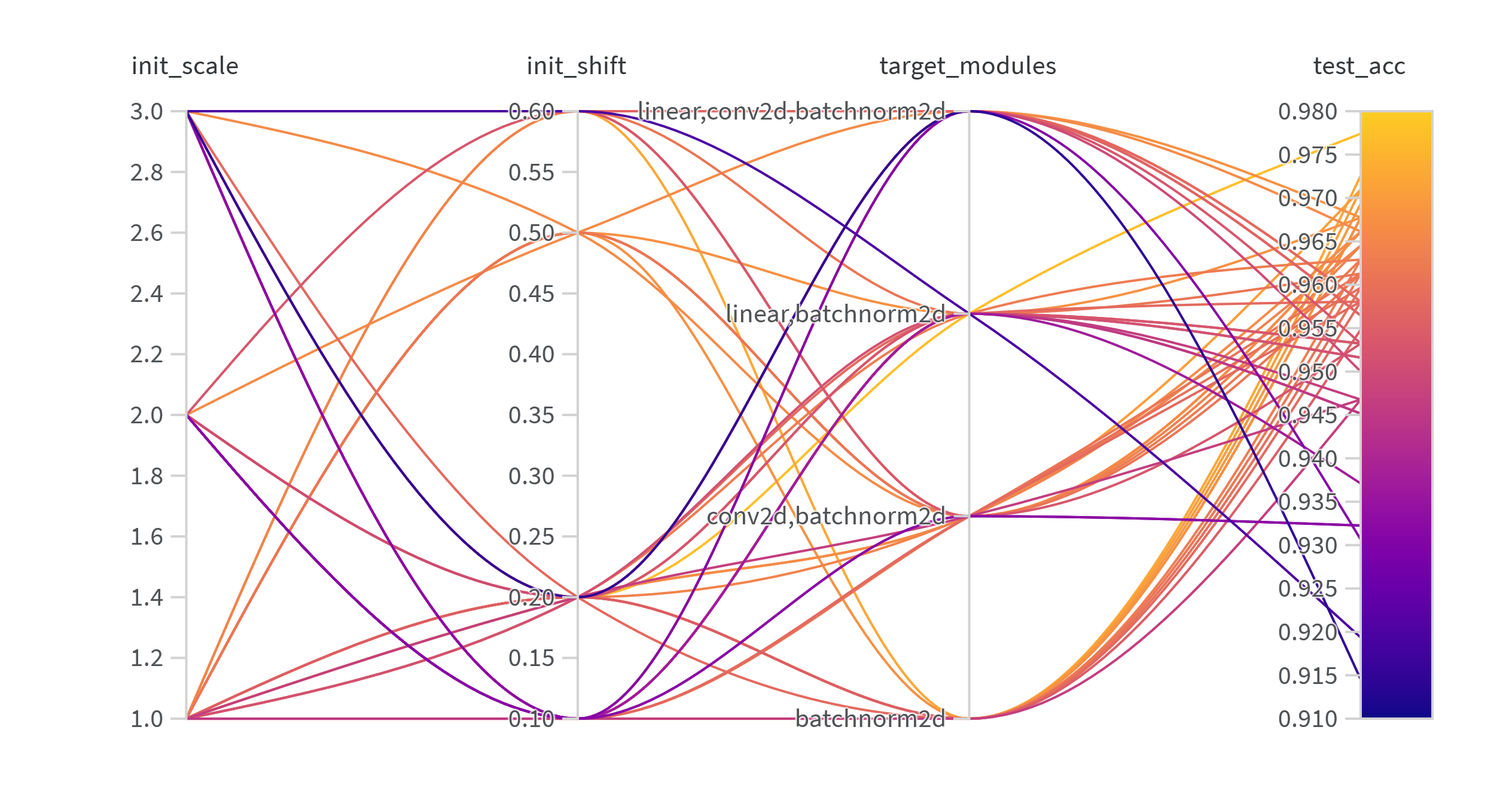}
\end{figure}

For example, as seen in Figure \ref{fig:sweep}

\subsection{5-fold Cross Validation}
Our final Results and metrics are derived from our 5-fold Cross Validation runs. For our purposes we used both validation and inference splits in the cross validation. Opting for 5-fold cross validation. A fixed 10\% of the data is used for inference at the end of a fold’s model training. The other 90\% is used in the Training and testing the model’s performance. Where 70\% of the data for training and the remaining 20\% for validation (testing). The model’s performance is then averaged over the 5 fold’s iterations and presented in table \ref{tab:kfold-results}

The number of training augmentations is described as "Augs" in the model column. The highest validated training regimen used was EfficientNet-b0 with 3 augs with full fine-tuning. Runners up, included other b0 and EfficientNet-b7 (3 Augs) regimines and AST (3 Augs) using SSF. Notably the custom CNN performed noticeably worse than all other methods used.

These results are a shift from our previous findings \cite{b1}, where the custom CNN was higher performing than the AST model. Interestingly, the EfficientNet B0 outperformed the former two, while only having roughly 4 million parameters.
\begin{table*}[ht]
\centering
\scriptsize  
\caption{5-Fold Cross Validation Results}
\label{tab:kfold-results}
\begin{tabularx}{\textwidth}{lXXXXXX}
\hline
\textbf{Model} & \textbf{Full} & \textbf{Classifier} & \textbf{Batchnorm} & \textbf{SSF} & \textbf{Ia3} & \textbf{OFT} \\
\hline
AST 0 Augs & 86.49\% $\pm$ 1.61\% & 88.24\% $\pm$ 2.01\% & - & 94.30\% $\pm$ 1.05\% & 92.76\% $\pm$ 0.43\% & 93.58\% $\pm$ 0.96\% \\
AST 3 Augs & 82.62\% $\pm$ 4.03\% & 88.49\% $\pm$ 0.55\% & - & \textbf{95.27\% $\pm$ 1.07\%} & 93.48\% $\pm$ 0.77\% & 93.87\% $\pm$ 1.90\% \\
Our CNN 0 Augs & 92.58\% $\pm$ 1.18\% & - & - & - & - & - \\
Our CNN 3 Augs & 94.73\% $\pm$ 0.82\% & - & - & - & - & - \\
ResNet-18 0 Augs & 94.41\% $\pm$ 0.82\% & 78.60\% $\pm$ 1.41\% & 89.82\% $\pm$ 0.52\% & 89.43\% $\pm$ 1.25\% & - & - \\
ResNet-18 3 Augs & 94.09\% $\pm$ 2.24\% & 76.31\% $\pm$ 1.83\% & 94.84\% $\pm$ 0.93\% & 94.59\% $\pm$ 0.47\% & - & - \\
ResNet-152 0 Augs & 92.47\% $\pm$ 1.47\% & 63.66\% $\pm$ 3.00\% & 57.56\% $\pm$ 30.34\% & 88.89\% $\pm$ 1.38\% & - & - \\
ResNet-152 3 Augs & 92.62\% $\pm$ 1.74\% & 65.02\% $\pm$ 2.56\% & 93.12\% $\pm$ 1.20\% & 94.80\% $\pm$ 1.29\% & - & - \\
MobileNet-Small 0 Augs & 92.80\% $\pm$ 1.42\% & 74.84\% $\pm$ 4.21\% & 83.19\% $\pm$ 1.99\% & 83.05\% $\pm$ 3.12\% & - & - \\
MobileNet-Small 3 Augs & 95.02\% $\pm$ 0.54\% & 74.12\% $\pm$ 1.45\% & 90.97\% $\pm$ 0.56\% & 86.81\% $\pm$ 3.84\% & - & - \\
MobileNet-Large 0 Augs & 92.37\% $\pm$ 1.25\% & 73.37\% $\pm$ 2.46\% & 77.24\% $\pm$ 2.13\% & 83.91\% $\pm$ 0.73\% & - & - \\
MobileNet-Large 3 Augs & 95.05\% $\pm$ 0.81\% & 71.94\% $\pm$ 1.81\% & 93.41\% $\pm$ 0.81\% & 90.97\% $\pm$ 1.78\% & - & - \\
EfficientNet-B0 0 Augs & 94.27\% $\pm$ 1.99\% & 94.27\% $\pm$ 1.99\% & 94.27\% $\pm$ 1.99\% & 94.27\% $\pm$ 1.99\% & - & - \\
EfficientNet-B0 3 Augs & \underline{\textbf{95.95\% $\pm$ 0.61\%}} & \textbf{95.81\% $\pm$ 0.56\%} & \textbf{95.91\% $\pm$ 1.01\%} & \textbf{95.77\% $\pm$ 0.97\%} & - & - \\
EfficientNet-B7 0 Augs & 93.58\% $\pm$ 0.50\% & 93.58\% $\pm$ 0.50\% & 93.58\% $\pm$ 0.50\% & 93.58\% $\pm$ 0.50\% & - & - \\
EfficientNet-B7 3 Augs & \textbf{95.66\% $\pm$ 1.07\%} & 94.87\% $\pm$ 1.35\% & 95.02\% $\pm$ 1.84\% & 95.45\% $\pm$ 0.61\% & - & - \\
\hline
\end{tabularx}
\end{table*}

\section{Conclusion}
In this paper we continued our work on small data regimens for deep UAV audio classification. We expanded the approach of using pre-trained, PEFT methods, and data augmentation to learn on our custom UAV dataset. 

Through empirical analysis we find that the highest validation accuracy leveraged the pre-trained EfficientNet-b0 model full fine-tuning along with 3 folds of training augmentations. The only other models to come close to achieving this performance were AST (3 augs) with SSF fine-tuning and EfficientNet-b7. 

In future works we would like to continue to scale up the audio dataset, as well as introduce visual and radar telemetry into the system. Further, we would like to explore into the theoretical workings of why the pre-training parameter efficient fine-tuning training paradigm is effective.

Of course it seems to be an obvious statement, but bigger is not always better. While, we have shown how to work around this limitation. It stands that smaller focused models can outperform on specific dataset scales.




\begin{thebibliography}{00}
\bibitem{b1} A. P. Berg, M. Y. Wang, Q. Zhang "4,500 Seconds: Small data training approaches for deep UAV audio classification," Proceedings 14th Int. Conf. Data Sci., Technol. Appl. (DATA), 2025. [Online] \url{https://www.insticc.org/Primoris/Resources/PaperPdf.ashx?idPaper=ScI3f83+ah4=}

\bibitem{b2} K. Zaman, M. Sah, C. Direkoglu, and M. Unoki, "A survey of audio classification using deep learning," IEEE Access, vol. 11, pp. 106620–106649, 2023, doi: 10.1109/ACCESS.2023.3318015.
\bibitem{b3} A. F. R. Nogueira, H. S. Oliveira, J. J. M. Machado, and J. M. R. S. Tavares, "Transformers for urban sound classification—A comprehensive performance evaluation," *Sensors*, vol. 22, no. 22, p. 8874, 2022, doi: 10.3390/s22228874.

\bibitem{b4}A. Berg, *UAV Classification* [Computer software], Available: \url{https://github.com/AndrewPBerg/UAV_Classification}.

\bibitem{b5} A. Berg, Weights \& Biases Project Logs [Online] Available: \url{https://wandb.ai/andberg9-self/projects}.
\bibitem{b6} A. Paszke, S. Gross, F. Massa, A. Lerer, J. Bradbury, G. Chanan, T. Killeen, Z. Lin, N. Gimelshein, L. Antiga, A. Desmaison, A. Köpf, E. Yang, Z. DeVito, M. Raison, A. Tejani, S. Chilamkurthy, B. Steiner, L. Fang, J. Bai, and S. Chintala, "PyTorch: An imperative style, high-performance deep learning library," arXiv preprint arXiv:1912.01703, Dec. 2019. [Online]. Available: \url{https://arxiv.org/abs/1912.01703}
\bibitem{b7} PyTorch Foundation, PyTorch Documentation, 2025. [Online]. Available: \url{https://docs.pytorch.org/docs/stable/index.html}.

\bibitem{b8} Hugging Face, Transformers Documentation, 2025 [Online]. Available: \url{https://huggingface.co/docs/transformers/en/index}
\bibitem{b9}Hugging Face, PEFT Documentation, 2025 [Online]. Available: \url{https://huggingface.co/docs/peft/en/index}
\bibitem{b10} Lightning AI, Pytroch Lightning Documentation, 2025 [Online]. Available: \url{https://lightning.ai/docs/pytorch/stable/}
\bibitem{b11} Weights \& Biases. Weights \& Biases Documentation, 2025 [Online]. Available: \url{https://docs.wandb.ai/}
\bibitem{b12} Docker. Docker Documentation, 2025 [Online]. Available: \url{https://docs.docker.com/}
\bibitem{b13} PyTorch Foundation, TorchAudio Documentation, 2025. [Online]. Available: 
\url{https://docs.pytorch.org/audio/stable/index.html}
\bibitem{b14} Audiomentations Documentation, 2025. [Online]. Available: \url{https://iver56.github.io/audiomentations/}
\bibitem{b15} TorchMetrics Documentation, 2025. [Online]. Available: \url{https://lightning.ai/docs/torchmetrics/stable/}
\bibitem{b16} Matplotlib Documentation, 2025. [Online]. Available: \url{https://matplotlib.org/stable/index.html} 
\bibitem{b17} scikit-learn Documentation, 2025. [Online]. Available: \url{https://scikit-learn.org/stable/} 
\bibitem{b18} NumPy Documentation, 2025. [Online]. Available: \url{https://numpy.org/doc/}
\bibitem{b19} Librosa Documentation, 2025. [Online]. Available: \url{https://librosa.org/doc/latest/index.html}
\bibitem{b20} S. S. Stevens, J. Volkmann, and E. B. Newman, "A scale for the measurement of the psychological magnitude pitch," *J. Acoust. Soc. Am.*, vol. 8, no. 3, pp. 185–190, 1937. [Online]. Available:   \url{https://pubs.aip.org/asa/jasa/article/8/3/185/673917/A-Scale-for-the-Measurement-of-the-Psychological}

\bibitem{b21} F. Wolf-Monheim, "Spectral and rhythm features for audio classification with deep convolutional neural networks," *arXiv preprint arXiv:2410.06927*, Oct. 2024. [Online]. Available: \url{https://arxiv.org/abs/2410.06927}

\bibitem{b22} Y. Wang, Z. Chu, I. Ku, E. C. Smith, and E. T. Matson, "A large-scale UAV audio dataset and audio-based UAV classification using CNN," in *Proc. 2022 Sixth IEEE Int. Conf. Robotic Comput. (IRC)*, Naples, Italy, Dec. 2022, pp. 186–189, doi: 10.1109/IRC55401.2022.00039.

\bibitem{b23} S. Kümmritz, "The sound of surveillance: Enhancing machine learning-driven drone detection with advanced acoustic augmentation," *Drones*, vol. 8, no. 3, p. 105, Mar. 2024. [Online]. Available: \url{https://doi.org/10.3390/drones8030105}

\bibitem{b24} A. P. Berg. Augmentations Colab Notebook, 2025. [Online]\url{https://colab.research.google.com/drive/1bl4RTQd7ENnMYEc4thwBwtocF-q1NYp2?usp=sharing]}

\bibitem{b25} M. Iman, H. R. Arabnia, and K. Rasheed, "A review of deep transfer learning and recent advancements," *Technologies*, vol. 11, no. 2, p. 40, Feb. 2023. [Online]. Available: \url{https://doi.org/10.3390/technologies11020040}

\bibitem{b26}  A. Mao, M. Mohri, and Y. Zhong, "Cross-entropy loss functions: Theoretical analysis and applications," arXiv preprint arXiv:2304.07288, Apr. 2023. [Online]. Available: \url{https://arxiv.org/abs/2304.07288}
\bibitem{b27} D. P. Kingma and J. Ba, "Adam: A method for stochastic optimization," arXiv preprint arXiv:1412.6980, Dec. 2014. [Online]. Available: \url{https://arxiv.org/abs/1412.6980}
\bibitem{b28} I. Loshchilov and F. Hutter, "Decoupled weight decay regularization," arXiv preprint arXiv:1711.05101, Nov. 2017. [Online]. \url{https://arxiv.org/abs/1711.05101}
\bibitem{b29} G. Alain and Y. Bengio, "Understanding intermediate layers using linear classifier probes," *arXiv preprint arXiv:1610.01644*, Nov. 2018. [Online]. Available: \url{https://arxiv.org/abs/1610.01644}
\bibitem{b30} Z. Qiu, W. Liu, H. Feng, Y. Xue, Y. Feng, Z. Liu, D. Zhang, A. Weller, and B. Schölkopf, "Controlling text-to-image diffusion by orthogonal finetuning," arXiv preprint arXiv:2306.07280, Jun. 2023. [Online]. Available: \url{https://arxiv.org/abs/2306.07280}
\bibitem{b31} H. Liu, D. Tam, M. Muqeeth, J. Mohta, T. Huang, M. Bansal, and C. Raffel, "Few-shot parameter-efficient fine-tuning is better and cheaper than in-context learning," arXiv preprint arXiv:2205.05638, May 2022. [Online]. Available: \url{https://arxiv.org/abs/2205.05638}
\bibitem{b32} D. Lian, D. Zhou, J. Feng, and X. Wang, "Scaling \& shifting your features: A new baseline for efficient model tuning," arXiv preprint arXiv:2210.08823, Oct. 2022. [Online]. Available: \url{https://arxiv.org/abs/2210.08823}
\bibitem{b33} A. P. Berg. UAV Classification SSF PEFT code, 2025 [Online]\url{https://github.com/AndrewPBerg/UAV_Classification/blob/master/src/models/ssf_adapter.py}
\bibitem{b34}  J. Frankle, D. J. Schwab, and A. S. Morcos, "Training BatchNorm and only BatchNorm: On the expressive power of random features in CNNs," arXiv preprint arXiv:2003.00152, Mar. 2020. [Online]. \url{Available: https://arxiv.org/abs/2003.00152}
\bibitem{b35} A. Dosovitskiy et al., "An image is worth 16x16 words: Transformers for image recognition at scale," arXiv preprint arXiv:2010.11929, Oct. 2020. [Online]. Available: \url{https://arxiv.org/abs/2010.11929}
\bibitem{b36} K. Palanisamy, D. Singhania, and A. Yao, "Rethinking CNN models for audio classification," arXiv preprint arXiv:2007.11154, Jul. 2020. [Online]. Available: \url{https://arxiv.org/abs/2007.11154}
\bibitem{b37} Y. Gong, Y.-A. Chung, and J. Glass, "AST: Audio Spectrogram Transformer," arXiv preprint arXiv:2104.01778, Apr. 2021. [Online]. Available: \url{https://arxiv.org/abs/2104.01778}
\bibitem{b38} O. Russakovsky et al., "ImageNet large scale visual recognition challenge," arXiv preprint arXiv:1409.0575, Sep. 2014. [Online]. Available: \url{https://arxiv.org/abs/1409.0575}
\bibitem{b39}  J. F. Gemmeke et al., "Audio Set: An ontology and human-labeled dataset for audio events," in *Proc. 2017 IEEE International Conference on Acoustics, Speech and Signal Processing (ICASSP)*, New Orleans, LA, USA, 2017, pp. 776–780, doi: 10.1109/ICASSP.2017.7952261. [Online]. Available: \url{https://ieeexplore.ieee.org/document/7952261}
\bibitem{b40} K. He, X. Zhang, S. Ren, and J. Sun, "Deep residual learning for image recognition," arXiv preprint arXiv:1512.03385, Dec. 2015. [Online]. Available: \url{https://arxiv.org/abs/1512.03385}

\bibitem{b41}
TorchVision, "ResNet — TorchVision main documentation," [Online]. Available: \url{https://pytorch.org/vision/main/models/resnet.html}
\bibitem{b42}
A. G. Howard et al., "MobileNets: Efficient convolutional neural networks for mobile vision applications," arXiv preprint arXiv:1704.04861, Apr. 2017. [Online]. Available: \url{https://arxiv.org/abs/1704.04861}
\bibitem{b43} A. Howard et al., "Searching for MobileNetV3," arXiv preprint arXiv:1905.02244, May 2019. [Online]. Available: \url{https://arxiv.org/abs/1905.02244}
\bibitem{b44}  TorchVision, "MobileNet V3 — TorchVision main documentation," [Online]. Available: \url{https://pytorch.org/vision/main/models/mobilenetv3.html}
\bibitem{b45}
M. Tan and Q. V. Le, "EfficientNet: Rethinking model scaling for convolutional neural networks," arXiv preprint arXiv:1905.11946, May 2019. [Online]. Available: \url{https://arxiv.org/abs/1905.11946}
\bibitem{b46} TorchVision, "EfficientNet — TorchVision main documentation," [Online]. Available: \url{https://pytorch.org/vision/main/models/efficientnet.html}
\bibitem{b47} V. Semenyuk, I. Kurmashev, A. Lupidi, D. Alyoshin, L. Kurmasheva, and A. Cantelli-Forti, "Advance and Refinement: The Evolution of UAV Detection and Classification Technologies," arXiv preprint arXiv:2409.05985, Sep. 2024. [Online]. Available: \url{https://arxiv.org/abs/2409.05985}
\bibitem{b48} Md. M. Rahman, S. Siddique, M. Kamal, R. H. Rifat, and K. D. Gupta, "UAV (Unmanned Aerial Vehicles): Diverse Applications of UAV Datasets in Segmentation, Classification, Detection, and Tracking," arXiv preprint arXiv:2409.03245, Sep. 2024. [Online]. Available: \url{https://arxiv.org/abs/2409.03245}
\bibitem{b49} V. Lialin, V. Deshpande, X. Yao, and A. Rumshisky, "Scaling Down to Scale Up: A Guide to Parameter-Efficient Fine-Tuning," arXiv preprint arXiv:2303.15647, Nov. 2024. [Online]. Available: \url{https://arxiv.org/abs/2303.15647}
\bibitem{b50} M. Y. Wang, D. C. Ramirez, E. Noonan, M. Linn, and Q. Zhang, "A Comprehensive Dataset and Visualization Tool for Drone Acoustic Signatures," in *Proc. 2024 Artificial Intelligence x Humanities, Education, and Art (AIxHEART)*, Laguna Hills, CA, USA, Sep. 2024, pp. 13–17, doi: 10.1109/AIxHeart62327.2024.00009. [Online]. Available: \url{https://www.researchgate.net/publication/388017272_A_Comprehensive_Dataset_and_Visualization_Tool_for_Drone_Acoustic_Signatures}
\bibitem{b51} M. Y. Wang, Z. Chu, I. Ku, E. C. Smith, and E. T. Matson, "A 15-Category Audio Dataset for Drones and an Audio-Based UAV Classification Using Machine Learning," *International Journal of Semantic Computing*, vol. 18, no. 2, pp. 257–272, 2024, doi: 10.1142/S1793351X24300048. [Online]. Available: \url{https://doi.org/10.1142/S1793351X24300048}



\end{thebibliography}
\end{document}